\documentclass[runningheads]{llncs}
\usepackage[T1]{fontenc}
\usepackage{graphicx}
\usepackage{booktabs}
\usepackage[misc]{ifsym}
\newcommand{\corr}{(\Letter)}
\usepackage{mwe}
\usepackage{url}
\usepackage{breakurl}
\usepackage[breaklinks]{hyperref}
\usepackage{algorithm}
\usepackage{algpseudocode}
\usepackage{mathtools}

\newcommand{\ours}{AHER}%

\newcommand{\citet}[1]{\cite{#1}}

\begin{document}

\title{Adaptable Hindsight Experience Replay for Search-Based Learning}

\author{Alexandros Vazaios\inst{1} \orcidID{0009-0007-2231-5694}\and
Jannis Brugger\inst{1,2} \orcidID{0000-0002-7919-4789} \corr \and
Cedric Derstroff\inst{1,2} \orcidID{0000-0002-7475-7546} \and
Kristian Kersting\inst{1,2,3,4} \orcidID{0000-0002-2873-9152} \and
Mira Mezini\inst{1,2,5}\orcidID{0000-0001-6563-7537} 
}

\authorrunning{A. Vazaios et al.}

\institute{Technical University of Darmstadt, Darmstadt, Germany \email{alexandros.vazaios@stud.tu-darmstadt.de}
\email{\{jannis.brugger, cedric.derstroff, mira.mezini, kristian.kersting\}@tu-darmstadt.de}
\and
Hessian Center for Artificial Intelligence (hessian.AI), Darmstadt, Germany
\and
German Research Center for Artificial Intelligence (DFKI), Germany
\and
Centre for Cognitive Science, Darmstadt, Germany
\and
National Research Center for Applied Cybersecurity ATHENE, Germany
}

\maketitle              %

\begin{abstract}
AlphaZero-like Monte Carlo Tree Search systems, originally introduced for two-player games, dynamically balance exploration and exploitation using neural network guidance. This combination makes them also suitable for classical search problems. However, the original method of training the network with simulation results is limited in sparse reward settings, especially in the early stages, where the network cannot yet give guidance. Hindsight Experience Replay (HER) addresses this issue by relabeling unsuccessful trajectories from the search tree as supervised learning signals.
We introduce
Adaptable HER (\ours{}), a flexible framework that integrates HER with AlphaZero, allowing easy adjustments to HER properties such as relabeled goals, policy targets, and trajectory selection. Our experiments, including equation discovery, show that the possibility of modifying HER is beneficial and surpasses the performance of pure supervised or reinforcement learning.

\end{abstract}

\section{Introduction}

Adaptive learning systems often face significant challenges due to the scarcity of high-quality training data in complex domains. To steer the search process in complex environments, neural-guided Monte Carlo Tree Search (MCTS) has emerged as a promising approach.
However, sparse learning signals can still result in ineffective and resource-heavy training.
To counter this problem, Hindsight Experience Replay (HER) transforms failures into valuable learning experiences by copying a complete historic trajectory and exchanging the original goal with one of the states in the search tree. Creating artificial positive training signals enables the neural network to capture a more comprehensive understanding of the search space. 
While other works \cite{goal-dir,learning-math} have already combined HER with neural-guided MCTS, we introduce a unifying, widely applicable
Adaptable HER framework (\ours{}) and extract the critical parts of HER. 
Our contributions are threefold: \textbf{I} We develop a \textbf{flexible HER framework} that can be seamlessly integrated into various environments and search algorithms by following the gymnasium interface \cite{gymnasium}. \textbf{II} We confirm that HER consistently improves the performance of neural-guided MCTS, but highlight the \textbf{need for a customizable HER framework}, as optimal configurations vary across environments. \textbf{III} We showcase in the domain of equation discovery that AlphaZero \cite{silver2018alphazero} with \textbf{HER outperforms both pure reinforcement learning and pure supervised learning}.

\section{Methodology}
Before explaining \ours{} in detail, we first give a brief overview of the AlphaZero training. We then finish this section with empirical results.

\subsection{AlphaZero}

The baseline learning approach selected for our experiments is a variation of AlphaZero \cite{silver2018alphazero}: a well-known neural-guided MCTS algorithm created by DeepMind that achieved superhuman performance in Shogi, Chess, and Go.
Its core idea lies in combining MCTS and a neural network policy.
The policy guides the MCTS, and the search results function as training targets.
After each episode, the network receives feedback by being trained on state transitions sampled from its Experience Replay. This paradigm can also be generalized to other games, including single-player ones, as required for our case. The Training routine is illustrated in the dashed frame of Fig. \ref{arch} as part of our \ours{} architecture.

\subsection{Adaptable Hindsight Experience Replay}
Depending on the task, there are various configurations for HER \cite{goal-dir,learning-math,her}, which we unify into the following four properties: (1) In the goal selection strategy, we can choose between the ``future'' strategy, relabeling states visited later in the current episode as HER-goals, or the ``final'' strategy, relabeling only the terminal state. (2) By selecting between a single- or multi-trajectory, we choose whether only the played trajectory of an episode is used to generate HER samples or a random subset from the MCTS search tree. (3) The following property defines how many HER samples are added to the replay buffer per trajectory. (4) The last property defines the policy learning target for the HER samples. We can choose between the original MCTS probabilities, a one-hot array, or a one-hot array with uniform noise. 

\begin{figure}
  \centering
  \makebox[\textwidth][c]{\includegraphics[width=1.05\textwidth]{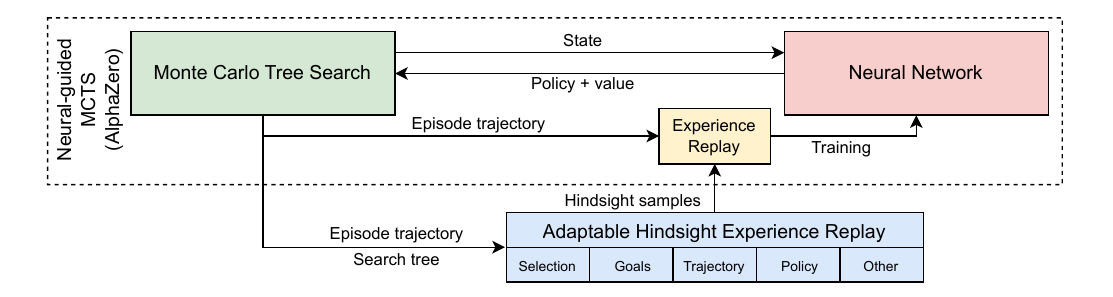}}
  \caption{Architectural overview of our learning setup. \ours{} is connected to the AlphaZero training loop. An iteration of the loop consists of performing neural-guided MCTS, sampling batches from the experience replay buffer, and training the predictor neural network.}
  \label{arch}
\end{figure}

By covering these four properties, we introduce a customizable implementation of HER for AlphaZero (Fig. \ref{arch}), which allows us to measure the influence of these properties on a set of learning domains.

\subsection{Setup}

Our implementation is
derived from the AlphaZero General \cite{a0general} and MuZero \cite{mu0} codebases with MCTS modules written by \citet{amex} and adapted to be compatible with single-player games by \citet{nged}. We chose to use three learning environments for our experiments: bit-flipping \cite{her}, point maze from Gymnasium Robotics \cite{gymnasium-r} (with action space discretized to 8 equidistant moves), and equation discovery from a table of measurements by \citet{nged}. The code used for this paper is available at: \url{https://github.com/my-curiosity/aher}.

For the policy, we utilize a 3-layer multi-layer perceptron (MLP) with 256, 256, and 128 neurons respectively,  ReLU activations, and a 0.3 dropout rate in bit-flipping and point maze. In equation discovery, the equation syntax trees and measurement datasets are separately encoded by a transformer (2 layers, 4 heads, 8-dimensional vectors) and an LSTM network (64 units), respectively. Their output is then combined, and separate MLPs (64/64, ReLU) are used to predict the value of the current state and action (rule of the grammar, describing how to construct an equation), to be applied next.

To evaluate performance in bit-flipping and point maze, we consider the success rate, mean total return, and the number of iterations necessary to reach a success rate of 80\%. In equation discovery, however, we are more interested in finding the best equations than in average performance. Therefore, we instead measure how many nodes are expanded by MCTS before a suitable formula is found, as done in previous work \cite{nged}. %

\subsection{Empirical Results}

\subsubsection{Bit-Flipping.} Bit-flipping is a task of transforming binary arrays by inverting values, one at a time, to reach given target configurations. It was used to demonstrate the effectiveness of HER to mitigate reward sparsity \cite{her}. %
Learning with \ours{} was most successful with 4 ``future'' goals. 
Further increases resulted in reaching a success rate of 0.8 faster at the cost of slightly degrading the episode returns. Even though one-hot policy targets led to approximately 3.5\% more failures, adding weak uniform noise negated this effect and even slightly raised the observed returns. Multi-trajectory HER failed; most likely because the average MCTS trajectories were much shorter than those played. The ``final'' selection strategy was not effective due to the used reward function: with negative rewards received at each step, episode returns in relabeled trajectories were almost equivalent to the original ones and lacked the diversity required for training the neural network.
    
\subsubsection{Point Maze.} Point maze is a physics-based navigation task, in which the agent must lead a ball to a specified destination in a maze. We selected the medium maze layout for all the experiments.
Here, HER's behavior was generally similar to our observations from bit-flipping, with the following exceptions: the optimal number of ``future'' samples was 8, further increases slowed down reaching the 80\% solved episodes mark; one-hot and noisy policy targets having a strictly negative influence (8--9\% lower success rate); multi-trajectory HER still suffering from the same issues as in bit-flipping but achieving comparable results with 1 and 4 random MCTS trajectories.
    
\subsubsection{Equation Discovery.} The goal of equation discovery is to create equations from a specified context-free grammar that describe a dataset of point measurements.
Unlike point maze, equation discovery is easily compatible with both reinforcement (RL) and supervised learning (SL) methods due to optimal solutions for each episode being known before interacting with the environment. Moreover, as only terminal states correspond to complete equations, the ``future'' selection strategy is not applicable, which limits the utility of single-trajectory HER variants.
On the other hand, by sampling ``final'' goals from random MCTS trajectories, we improved over the previous best performance achieved by both plain 
RL and SL. However, the latter remains significantly less time-consuming. An optimum point was measured at 24 samples with the noise-free one-hot policy targets, as original MCTS probabilities are irrelevant outside the played trajectory.
 
\subsubsection{General Results.} 
\begin{figure}
  \centering
  \makebox[\textwidth][c]{\includegraphics[width=1.1\textwidth]{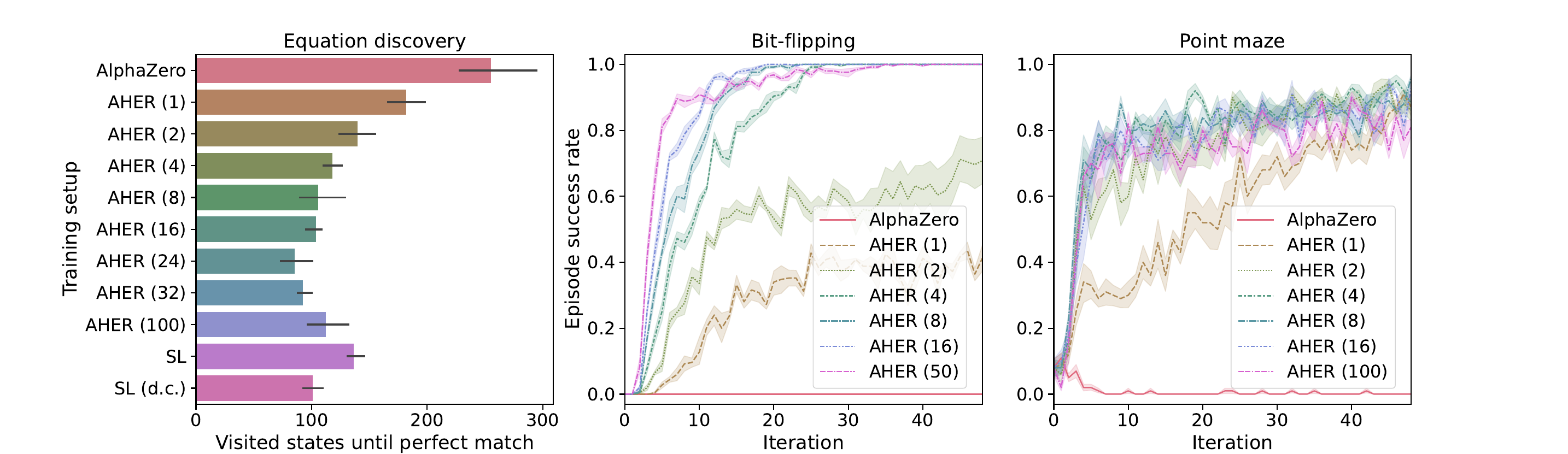}}
   \caption{Performance of AHER-augmented AlphaZero. (left) Visited states in the search tree until the correct equation is found. AlphaZero refers to training only with the probabilities from the MCTS, AHER($k$) for adding HER samples from $k$ trajectories, and SL (d.c.) for supervised learning with dataset changes.
   Dataset changes entail sampling new $x$ and fitting $y$ measurements after each training step to increase training diversity. We chose to also perform this during AHER goal relabeling.
   The HER samples are added with the ``final'' goal selection strategy and one-hot policy targets. Failed searches count as 1,000 visited states. (middle and right) Success rate for solving bit-flipping with 50 bits (middle) and point maze (right), depending on the number of HER samples. The HER samples are added with the ``future'' goal selection strategy from the played trajectory and original MCTS probabilities. We display the mean and 95\% confidence intervals of 5 runs.
   }
  \label{exp}
\end{figure}

The amount of HER samples used and, consequently, the ratio of original data to hindsight data in the experience replay affects AlphaZero's performance the strongest (cf. Fig. \ref{exp}). Tweaking this ratio is essential for ensuring successful training and can be seen as controlling the amount of supervision necessary for the task. With too few HER samples, the agent struggles to learn in sparse reward settings, but with too many,
it forgets past experiences and detaches from them, becoming less stable (as discussed by \citet{ris}). 
The data ratio of HER is mostly influenced by two properties, namely, the number of future goals and the number of trajectories used for relabeling.
Multi-trajectory HER is generally more difficult to apply as it performs poorly if the trajectories are short. Still, it remains a good solution for situations in which the ``future'' goal selection is unavailable, e.g., equation discovery. The tested policy targets demonstrated relatively similar results. However, one-hot encoding was naturally more appropriate when using random trajectories, as it reduces the amount of noise from the obsolete MCTS probabilities.
    
\subsubsection{Including 
Further
HER Advancements.} In addition to these results, we also adapted 3 HER improvement techniques to our setup: aggressive rewards \cite{archer}, experience ranking \cite{her-er}, and combined experience replay \cite{hcer}. As a result, a slight performance increase was reached in all compatible tasks. We see this direction of work as promising and having the potential to raise the effectiveness of neural-guided MCTS with HER without requiring the implementation of new concepts from scratch. 

\section{Related Work}
Although HER was initially intended for off-policy deep RL, it has been deployed with neural-guided MCTS in AlphaZeroHER \cite{goal-dir} and Minimo \cite{learning-math}. AlphaZeroHER shows that such a setup can outperform DQN (Deep Q-Network) with HER in multiple domains, but only considers the played trajectory for goal relabeling, ``future'' goal selection strategy, and original MCTS probabilities as a policy learning target for AlphaZero. Conversely, Minimo applies multi-trajectory HER to prove mathematical and logical conjectures, but utilizes environment-specific optimizations such as ignoring irrelevant subtrajectories. Additionally, the work uses a large language model agent, which changed the required form for training samples. Continuing this line of work, we analyze the influence of HER on learning in different configurations and on various tasks.

\section{Limitations and Outlook}
Due to the nature of MCTS, which is used as a search backbone, our implementation does not support continuous action spaces, and the necessary discretization is not always possible without a loss in performance or an increased training time. Furthermore, we only use straightforward policy learning targets (original probabilities and one-hot) that are generalizable to all domains. In future work, we would like to analyze how \ours{} can be used with more target-specific policies. Additionally, this work only considers deterministic state transition sequences. Systematically studying HER for probabilistic transitions and noisy states could further increase the application scenarios where HER improves the learning. Combining \ours{} with existing HER advancements is also expected to strengthen its performance; especially useful would be the curriculum mechanism \cite{cur-her}, which could aid both trajectory and goal choices for relabeling.

\section{Conclusion}

This work introduces \ours{}: an adaptable HER implementation for neural-guided AlphaZero-like MCTS, in which multiple architectural aspects can be configured. Leveraging it, we first analyze the influence of applying HER to AlphaZero to solve three domains.
HER's inclusion in the AlphaZero setup allows us to solve domains otherwise unsolvable by neural-guided MCTS. We observe that being able to select a type of HER to use with a given task can be beneficial, even for tasks such as point maze and bit-flipping, where the highest performance metrics correspond to the ``baseline'' variant by \citet{goal-dir}. Having the flexibility to change the HER properties allows us to measure the relative importance of each HER property for successful training. Moreover, it is essential for new domains, where the preferred type of HER-assisted learning is unknown and difficult to guess. 
In the equation discovery task, we  show that neural-guided MCTS with \ours{} can outperform pure supervised and reinforcement learning if configured correctly. %

\begin{credits}
\subsubsection*{\ackname}The authors gratefully acknowledge the computing time provided to them on the high-performance computer Lichtenberg II at TU Darmstadt, funded by the German Federal Ministry of Education and Research (BMBF) and the State of Hesse.

This research project was partly funded by the Hessian Ministry of Science and the Arts (HMWK) within the projects ``The Third Wave of Artificial Intelligence - 3AI'' and hessian.AI.

\subsubsection*{\discintname}
The authors have no relevant financial or non-financial interests to disclose.
\end{credits}
\bibliographystyle{splncs04}
\bibliography{bibliography}
\end{document}